\documentclass[conference]{IEEEtran}
\IEEEoverridecommandlockouts
\usepackage{cite}
\usepackage{amsmath,amssymb,amsfonts}
\usepackage{float, stackengine}
\usepackage{algorithmic}
\usepackage{graphicx}
\usepackage{textcomp}
\usepackage{xcolor}
\usepackage{booktabs}
\usepackage{multirow}
\usepackage[hyphens]{url}
\def\BibTeX{{\rm B\kern-.05em{\sc i\kern-.025em b}\kern-.08em
    T\kern-.1667em\lower.7ex\hbox{E}\kern-.125emX}}
\begin{document}

\title{Leveraging Large Language Models and Topic Modeling for Toxicity Classification \thanks{\\$^*$ contribute equally.\\Preprint submitted to CNC 2025.}}

\author{\IEEEauthorblockN{1\textsuperscript{st} Haniyeh Ehsani Oskouie$^*$}
\IEEEauthorblockA{\textit{Department of Computer Science} \\
\textit{University of California, Los Angeles}\\
Los Angeles, United States \\
haniyeh@cs.ucla.edu}
\and
\IEEEauthorblockN{2\textsuperscript{nd} Christina Chance$^*$}
\IEEEauthorblockA{\textit{Department of Computer Science} \\
\textit{University of California, Los Angeles}\\
Los Angeles, United States \\
cchance@ucla.edu}
\and
\IEEEauthorblockN{3\textsuperscript{rd} Claire Huang$^*$}
\IEEEauthorblockA{\textit{Department of Computer Science} \\
\textit{University of California, Los Angeles}\\
Los Angeles, United States \\
clairehuang1@ucla.edu}
\and
\IEEEauthorblockN{4\textsuperscript{th} Margaret Capetz$^*$}
\IEEEauthorblockA{\textit{Department of Computer Science} \\
\textit{University of California, Los Angeles}\\
Los Angeles, United States \\
mcapetz17@ucla.edu}
\and
\IEEEauthorblockN{5\textsuperscript{th} Elizabeth Eyeson$^*$}
\IEEEauthorblockA{\textit{Department of Computer Science} \\
\textit{University of California, Los Angeles}\\
Los Angeles, United States \\
eeyeson@ucla.edu} 
\and
\IEEEauthorblockN{6\textsuperscript{th} Majid Sarrafzadeh}
\IEEEauthorblockA{\textit{Department of Computer Science} \\
\textit{University of California, Los Angeles}\\
Los Angeles, United States \\
 majid@cs.ucla.edu}
}

\maketitle

\begin{abstract}
Content moderation and toxicity classification represent critical tasks with significant social implications. However, studies have shown that major classification models exhibit tendencies to magnify or reduce biases and potentially overlook or disadvantage certain marginalized groups within their classification processes. Researchers suggest that the positionality of annotators influences the gold standard labels in which the models learned from propagate annotators' bias. To further investigate the impact of annotator positionality, we delve into fine-tuning BERTweet and HateBERT on the dataset while using topic-modeling strategies for content moderation. The results indicate that fine-tuning the models on specific topics results in a notable improvement in the F1 score of the models when compared to the predictions generated by other prominent classification models such as GPT-4, PerspectiveAPI, and RewireAPI. These findings further reveal that the state-of-the-art large language models exhibit significant limitations in accurately detecting and interpreting text toxicity contrasted with earlier methodologies. Code is available at https://github.com/aheldis/Toxicity-Classification.git.
\end{abstract}


\section{Introduction}
Content moderation is important to mitigating the spread of potentially harmful content like hate speech, self-harm, or harassment on social media platforms. Without effective moderation, users risk being exposed to psychological harm or perpetuating harm itself. Thus, upholding civility, psychological safety and inclusivity in social media interactions depends upon robust content moderation mechanisms. This is important in our increasingly digital world. 

Popular toxicity classification and moderation techniques tend to rely on human annotations due to limitations in automated labeling. However, such annotations can amplify bias due to the identities of the annotators, lived experiences, societal / cultural norms and personal beliefs, a concept known as positionality of the annotator \cite{nlpositionality}. This subjectivity can inadvertently perpetuate stereotypes and marginalization in datasets and thus impact the performance of machine learning models. Therefore, investigating the behavior of neural networks using an unbiased dataset is fundamental to the development of reliable, fair, and effective AI systems. Additionally, with the growing recognition of large language models (LLMs), ensuring that they do not produce or amplify toxic content is crucial for user safety and platform integrity. So far, many researchers have shown the inability of these models to distinguish toxicity within text \cite{deshpande-etal-2023-toxicity}. Transfer learning may be used to improve content moderation systems by leveraging pre-trained models' knowledge. This approach allows for more efficient and effective toxicity classification by fine-tuning existing models on domain-specific data, potentially reducing bias and improving performance across diverse contexts. The objective of this paper is to introduce a topic-modeling-enhanced fine-tuning approach applied to toxicity data, with the aim of achieving superior results compared to existing toxicity classification models. 

\begin{table*}[h]
\footnotesize	
\centering
\caption{Topic distribution produced by Latent Dirichlet Allocation.}
\resizebox{0.55\textwidth}{!}{
\begin{tabular}{cll}
\toprule
\multicolumn{3}{c}{Number of Topics} \\ 
\multicolumn{1}{c}{3} & \multicolumn{1}{c}{6} & \multicolumn{1}{c}{10} \\ \midrule
\multicolumn{1}{l}{\begin{tabular}[c]{@{}l@{}}0 : woman people wa always never\\ 1 : people white get issue race\\ 2 : people muslim like make want\end{tabular}} & 
\multicolumn{1}{l}{\begin{tabular}[c]{@{}l@{}}0 : get gay love know make\\ 1 : people race issue make without\\ 2 : white wa men order would\\ 3 : people subhuman muslim white black\\ 4 : woman people always never take\\ 5 : people like wa ha even\end{tabular}} & 
\begin{tabular}[c]{@{}l@{}}0:  f*ck get people old white\\ 1 : people know muslim non thankfully\\ 2 : like would white black dwarf\\ 3 : history people without man black\\ 4 : people like white woman race\\ 5 : ret*rd make back people immigrant\\ 6  : woman always never take idiot\\ 7 : wa people ha time started\\ 8 : people wa really white million\\ 9 : woman men get people like\end{tabular} \\ \bottomrule
\end{tabular}
}
 
\label{tab:topic_modeling}
\end{table*}

\section{Background}
Large language models (LLMs), including the Generative Pre-trained Transformer (GPT) developed by OpenAI \cite{gpt} and the Bidirectional Encoder Representations from Transformers (BERT) developed by Google \cite{bert}, have shown great ability in understanding and generating human language.
These models are pre-trained on extensive amounts of data and are fine-tuned  and applied for specific tasks, such as content generation, translation, code development, sentiment analysis, and more \cite{radford2019language, mt5, few-shot, survey-code, sentiment} using both single-machine and federated learning approaches \cite{federated}. Due to the growing popularity of the LLMs, there has been an increasing concern about the performance of LLMs in understanding toxicity, which plays a crucial role in creating safer, more inclusive online spaces. Unfortunately, it has been shown that neural networks, especially LLMs, often exhibit biases caused by biases in their training data, reflecting the cultural and personal backgrounds of annotators \cite{nlpositionality}. Moreover, some studies indicate that significantly more toxic language can be generated using GPT by assigning its persona \cite{deshpande-etal-2023-toxicity}. While many studies have sought to enhance the performance of LLMs regarding toxicity detection \cite{efficient, hot, toxicgit}, none have examined their performance on toxicity datasets generated based on annotator positionalities.
These showcase the necessity of implementing an effective approach to address the toxicity present within texts while working with LLMs.

The use of short-form text, such as Tweets, has been a popular area of study as the limited context forces the model to find latent cues and patterns compared to longer text in which there is more context for classification. Many studies that utilize tweets for crisis analysis in a classification setting leverage the accessibility of tweets as well as its ability to capture cultural and social sentiment at any given time, especially used in the content moderation domain \cite{adhikari-etal-2022-covid, seeberger-riedhammer-2022-enhancing}.

Several works explore the use of author-pooled Latent Dirichlet Allocation (LDA) to extract discussion topics from Twitter data related to climate change \cite{Dahal}. Similarly, there has been a focus on comment moderation, utilizing a topic-aware model to enhance automatic moderation by incorporating semantic features from topic models \cite{zosa2021comments}. In a related context, various studies delve into enhancing word embeddings with topical information for toxic content detection, showcasing the effectiveness of incorporating topic-specific data in classification tasks \cite{kim}.

These studies collectively underscore the significance of toxicity classification and considering topic modeling, contextual factors, and specialized features in toxicity assessment and content moderation. Expanding on the knowledge gleaned from prior research, our objective in this study is to advance the field by utilizing a topic-modeling methodology for enhancing the performance in toxicity classification for short-form text. For this purpose, we first use LDA topic modeling \cite{lda} for topic clustering on data. We then fine-tune BERTweet \cite{bertweet} and HateBERT \cite{hatebert} on the subsets of the data generated by LDA to learn embeddings and neural representations that capture the key features influencing the classification of a toxic tweet. Ultimately, to evaluate the effectiveness of the proposed approach, we compare its results against those of existing toxicity detection models.

\begin{figure}[h!]
    \centering
    \includegraphics[width=0.5\textwidth]{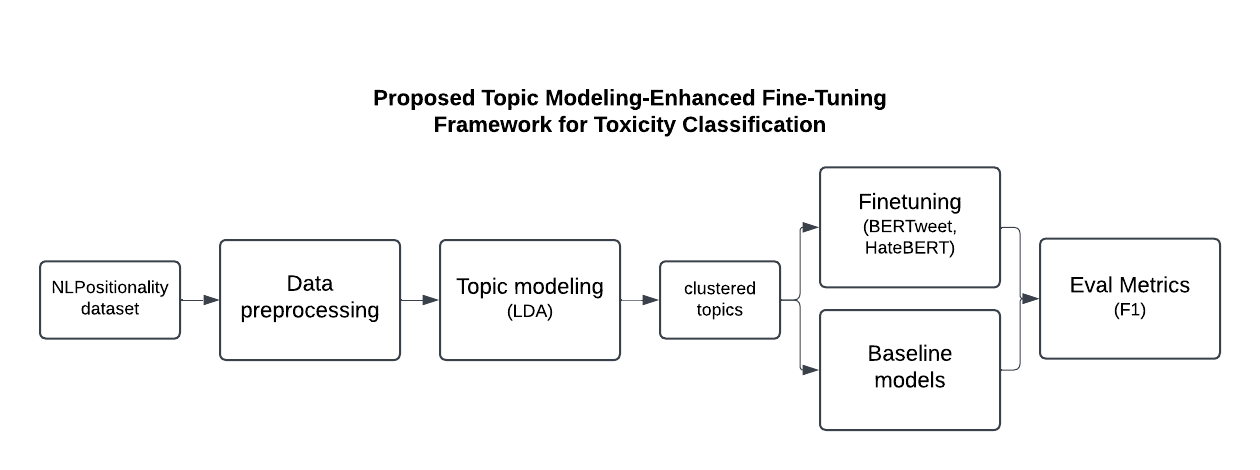} % Adjust the width as needed
    \caption{Diagram illustrating the proposed strategy for toxicity classification using topic modeling and fine-tuning.}
    \label{fig:toxicity_strategy}
\end{figure}

\section{Method}

An overview of our method is shown in Fig. \ref{fig:toxicity_strategy}. The details include:

\subsection{Dataset}
The dataset selected for our analysis is NLPositionality, a benchmark dataset consisting of labeled toxic tweets and annotator demographic metadata. This dataset is derived from \cite{santy-etal-2023-nlpositionality}, which introduces a framework for characterizing design biases and quantifying the positionality of natural language processing (NLP) datasets and models. By utilizing this dataset, we ensure that the analysis of toxicity in LLMs is accurate.

\subsection{Data preprocessing}
For preprocessing the data, we employ several techniques, including sentence tokenization, stop word removal, and lemmatization that enhance the quality of our input data. To handle the tokenization process for our models, we follow the same configuration settings that were established for the BERT models \cite{bertweet, hatebert}.

\subsection{Topic clustering}
Latent Dirichlet Analysis (LDA), a popular statistical technique \cite{lda}, is performed for topic modeling on the training data and then applied to the test set. Using this method, we cluster the toxicity dataset into 3, 6, and 10 topics. The examples of this clustering are shown in Table \ref{tab:topic_modeling}. Our findings demonstrate that performing topic modeling when the number of clusters $k$ is larger produces more insightful and expressive topics. Eventually, we use $k=3$ for fine-tuning the models as we have a smaller dataset. Further subsetting would produce insignificant numbers that we cannot draw assumptions.

\subsection{Models}
For toxicity classification, we utilize two pre-trained models including BERTweet and HateBERT. BERTweet is a model that was trained on a more general corpus of tweets, while HateBERT was trained on a more relevant corpus to our research: hate-speech related texts. As BERTweet was fine-tuned on short form tweets, the goal was to leverage the model's ability to perform a task on limited context. On the other hand, HateBERT  was obtained by fine-tuning the English BERT base uncased model on ToxiGen \cite{toxigen} data. The goal of this model was to leverage its task-specific context and its learned ability in understand implicit toxicity to be able to generalize on more explicit examples \cite{bertweet, hatebert}. These specifications make them suitable for fine-tuning with the purpose of toxicity analysis.

\begin{table*}[htbp]
  \centering
  \caption{F1 score for BERT models with different seeds.}
  \resizebox{0.55\textwidth}{!}{
  \label{tab:f1_scores}
  \begin{tabular}{l|l|ccccc|cc}
    \toprule
    \textbf{Model} & \textbf{Data split} & \textbf{Seed 0} & \textbf{Seed 1} & \textbf{Seed 2} & \textbf{Seed 3} & \textbf{Seed 9} & \textbf{Average} & \textbf{Stdev} \\
    \midrule
    \multirow{4}{*}{BERTweet} 
    & Topic 0 & 0.5588 & 0.5566 & 0.5566 & 0.5588 & 0.5588 & 0.5579 & 0.0012 \\
    & Topic 1 & 0.4778 & 0.4778 & 0.4778 & 0.4778 & 0.4778 & 0.4778 & 0.0000 \\
    & Topic 2 & 0.4659 & 0.4600 & 0.4659 & 0.4659 & 0.4600 & 0.4636 & 0.0032 \\
    & Full data & 0.4610 & 0.4631 & 0.4610 & 0.4560 & 0.4610 & 0.4604 & 0.0026 \\
    \midrule
    \multirow{4}{*}{HateBERT} 
    & Topic 0 & 0.5498	& 0.5498	& 0.5498	& 0.5498	& 0.5498 & 0.5498	& 0.0000 \\
    & Topic 1 &0.4767	& 0.4767	& 0.4767	& 0.4767	& 0.4767 & 0.4767	& 0.0000\\
    & Topic 2 & 0.4571	& 0.4572	& 0.4572	& 0.4572	& 0.4572 & 0.4572	& 0.0000 \\
    & Full data & 0.4831 & 0.4765 & 0.4837 & 0.4835 & 0.4852 & 0.4824 & 0.0034 \\
    \bottomrule
  \end{tabular}
}
\end{table*}

\subsection{Transfer learning}
The potential advantages of transfer learning include reducing the risk of overfitting by preserving the generalization ability of the pre-trained model, saving computational resources and time, and preventing catastrophic forgetting by preserving the features learned by the pre-trained model \cite{transfer}. This is particularly beneficial for HateBERT, as its pre-trained weights are already well-aligned with our task.  
Thus, we fine-tune the models for toxicity classification on various data splits. In this regard, we freeze all layers except for the classification head.
The hyperparameters used for tranfer learning are as follows: learning rate of $5\mathrm{e}{-5}$, $0$ warm up steps, and $70$ epochs. To ensure reliable results, each experiment is repeated five times using different manual seeds and both the mean and standard deviation of the outcomes are reported.

\section{Results}
Table \ref{tab:f1_scores} presents the F1 score of the BERT models for different seeds.
From the reported results, we see for BERTweet and HateBERT, fine-tuning the models on individual topics improved the F1 score compared to fine-tuning on the full dataset on average. The most significant improvement in the F1 score was for Topic 0, while the differences between the full dataset and Topic 1 and 2 were more marginal in comparison. One possible explanation may be that Topic 0 provides more distinct, consistent patterns of toxicity for the model to recognize, while Topics 1 and 2 may contain more varied, nuanced forms of toxicity. Interestingly, for HateBERT, the model fine-tuned on the entire dataset performed second to best, while for BERTweet, the model fine-tuned on the full dataset performed the worst.

% \begin{table}[htbp]
%   \centering
%   \caption{F1 score of other baselines for different data splits.}
%   \label{tab:others}
%   \begin{tabular}{l|ccccc}
%     \toprule
%     \textbf{Model} & \textbf{Topic 1} & \textbf{Topic 2} & \textbf{Topic 3} & \textbf{Full data} \\
%     \midrule
%     PerspectiveAPI & 0.3854 & 0.3857 & 0.3143 & 0.3636\\
%     RewireAPI & 0.4386 & 0.4153 & 0.4295 & 0.4278 \\
%     HateRoberta & 0.3689 & 0.4216 & 0.3310 & 0.3788 \\
%     GPT-4 & 0.3966 & 0.4163 & 0.3915 & 0.4054 \\
%     \bottomrule
%   \end{tabular}
% \end{table}

\begin{table}[htbp]
\centering
\caption{F1 score comparison between baselines and BERT models.}
\label{tab:others}
\resizebox{0.75\columnwidth}{!}{
\begin{tabular}{l|cccc}
\toprule
\textbf{Model} & \textbf{Topic 0} & \textbf{Topic 1} & \textbf{Topic 2} & \textbf{Full data} \\
\midrule
PerspectiveAPI & 0.3854 & 0.3857 & 0.3143 & 0.3636 \\
RewireAPI      & 0.4386 & 0.4153 & 0.4295 & 0.4278 \\
HateRoberta    & 0.3689 & 0.4216 & 0.3310 & 0.3788 \\
GPT-4          & 0.3966 & 0.4163 & 0.3915 & 0.4054 \\
\midrule
BERTweet       & 0.5579 & 0.4778 & 0.4636 & 0.4604 \\
HateBERT       & 0.5498 & 0.4767 & 0.4572 & 0.4824 \\
\bottomrule
\end{tabular}
}
\end{table}

Due to the absence of similar datasets for toxicity detection based on annotator positionalities, we did not compare our results with any other datasets.
Instead, we analyzed the performance of other baselines for toxicity detection on both the full data and its splits. These baselines include GPT-4 \cite{gpt}, PerspectiveAPI \cite{perspectiveapi}, RewireAPI \cite{rewire}, and HateRoberta without fine-tuning \cite{toxigen}. For all these models, we employed the similar settings specified by \cite{nlpositionality}. The results are demonstrated in Table \ref{tab:others}. As indicated, all of these models exhibit lower performance compared to our fine-tuned models. This highlights that current toxicity detection models and large language models like GPT-4 are not effectively trained to identify toxicity in text. This indicates that, similar to our approach, they require additional training or fine-tuning on NLPositionality-like datasets to enhance their robustness to toxicity.

\section{Discussion}
\subsection{Analysis}
Table \ref{tab:micro_f1_precision_recall} displays majority voting performed across seed runs to get predicted labels.  It appears that the breakdown by topic did not yield notable differences in performance compared to the full dataset, suggesting that there is no one cluster that capture more latent and semantic features and information that influence model prediction.

\begin{figure*}[h!]
    \centering
    \small
    \stackunder[2pt]{\includegraphics[width=0.3\textwidth]{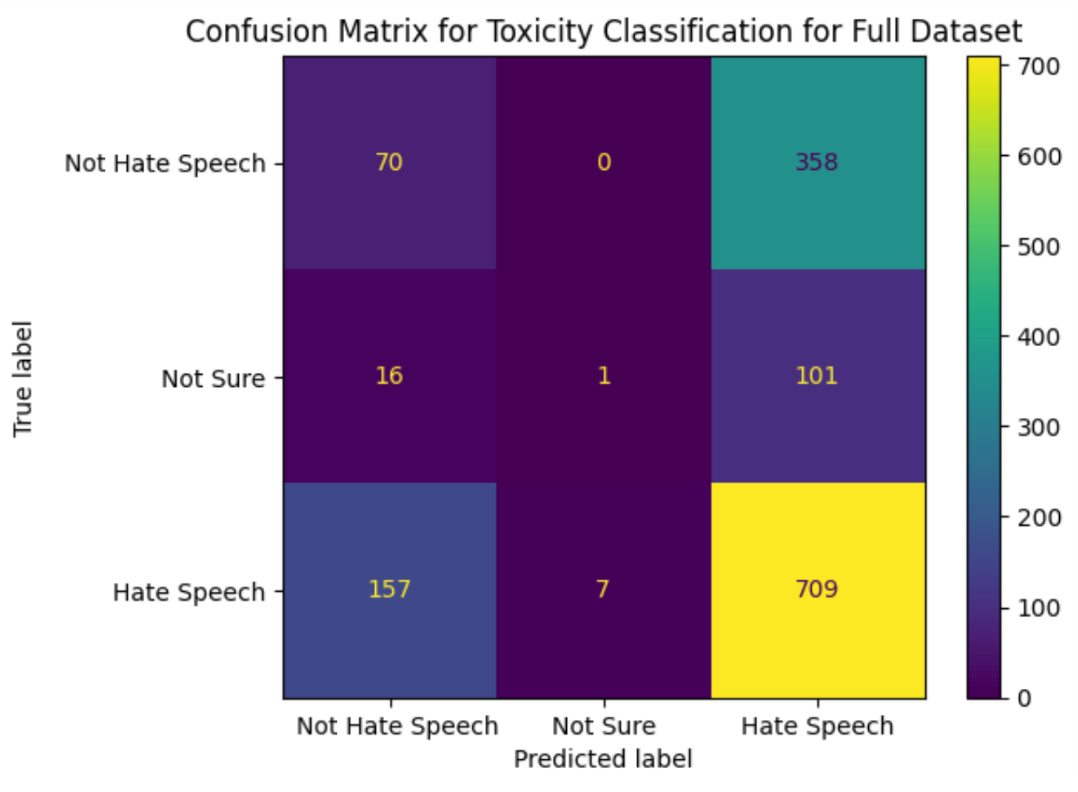}}{(a) BERTweet}
    \hspace{3cm}
    \stackunder[2pt]{\includegraphics[width=0.3\textwidth]{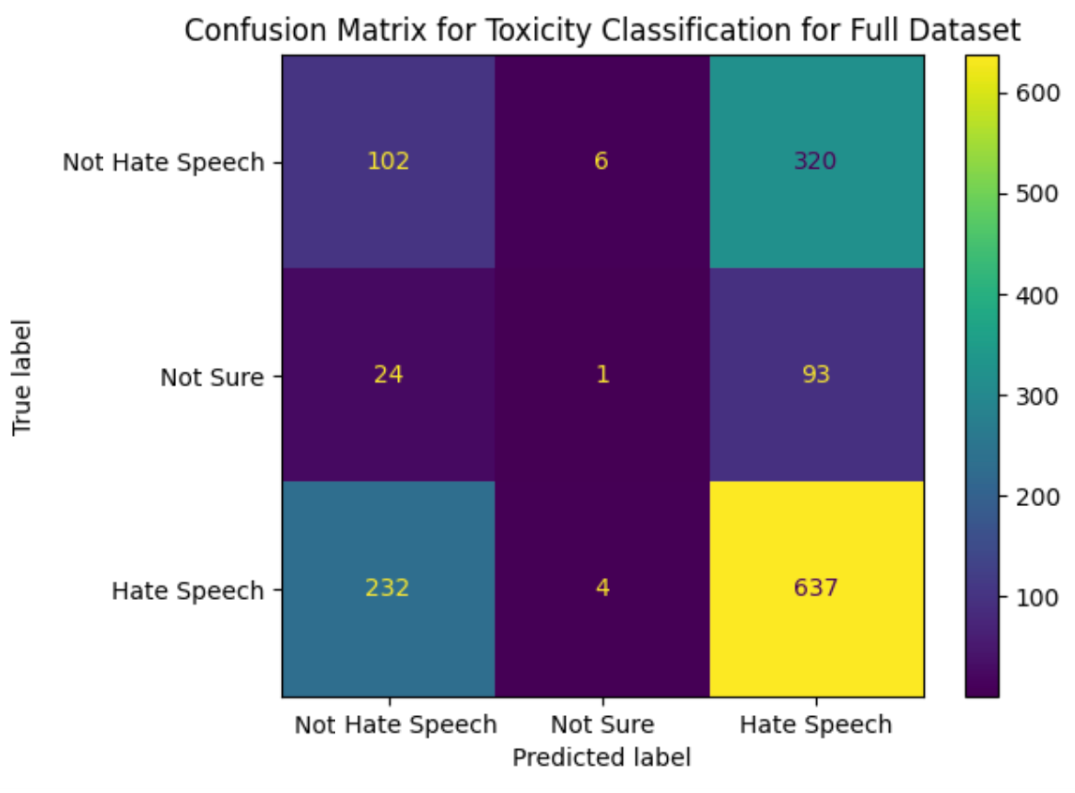}
    }{(b) HateBERT}
    \caption{
    Confusion matrices for fine-tuned BERTweet and HateBERT.
    }
    \label{Figure:confusion}
\end{figure*}

Data subsets were further grouped by gender and ethnicity and visualized using confusion matrices as illustrated in Fig. \ref{Figure:confusion}. Since positionality bias propagates through the form of incorrectly labeling examples as hate speech, we were interested in the true positive rate (TPR), true negative rate (TNR), false positive rates (FPR), as well as the precision and f1 score. As seen in Fig. \ref{Figure:confusion}, both fine-tuned BERTweet and HateBERT had high FPRs and TNRs. For Black annotators especially, BERTweet had a high TPR and recall overall.

\begin{table}[h]
 \centering
 \caption{Statistics for BERT models}
 \label{tab:micro_f1_precision_recall}
 \resizebox{0.75\columnwidth}{!}{
 \begin{tabular}{l|l|ccc}
   \toprule
   \textbf{Model} &
   \textbf{Data split} & \textbf{Micro F1} & \textbf{Precision} & \textbf{Recall} \\
   \midrule
   \multirow{4}{*}{BERTweet} & Full data & 0.5497 & 0.5497 & 0.5497 \\
   & Topic 0 & 0.5242 & 0.5242 & 0.5242 \\
   & Topic 1 & 0.5574 & 0.5574 & 0.5574 \\
   & Topic 2 & 0.5172 & 0.5172 & 0.5172 \\
   \midrule
   \multirow{4}{*}{HateBert} & Full data & 0.5215 & 0.5215 & 0.5215 \\
   & Topic 0 & 0.5196 & 0.5196 & 0.5196 \\
   & Topic 1 & 0.5410 & 0.5410 & 0.5410 \\
   & Topic 2 & 0.5011 & 0.5011 & 0.5011 \\
   \bottomrule
 \end{tabular}
 }
\end{table}

\subsection{Limitations}
Topic modeling was not as expressive as necessary due to the variety in tweets. With a large number of topics, we had a more understandable grouping, but due to the size of the dataset as well as the knowledge that further subsets of the data would be too small for analysis, we decided to use a smaller number of topics. In addition, BERTweet was pre-trained on general tweets, which may not be specific enough for our downstream task of training for toxicity classification. Further, because HateBERT was pre-trained on a binary toxicity classification dataset, the inclusion of a third label for our dataset during the fine-tune process may have contributed to the high error rates for that new label.

\section{Conclusion}
Our work was motivated by the fact that effective content moderation is critical to limit the spread of harmful content on social media platforms. We aimed to tackle the issue of biases introduced by human annotations in toxicity classification, which can be influenced by annotators' identities, experiences, and societal norms. Specifically, we explored the impact of fine-tuning BERTweet and HateBERT on topic-specific subsets of the NLPositionality dataset and its generalization to other platforms. We accomplished this by using topic modeling via LDA to find latent themes in toxic and non-toxic tweets. Our results demonstrate that fine-tuning the models on specific topics significantly enhances the F1 score compared to the other existing toxicity models. Future research should focus on mitigating the biases present in widely used models like GPT, as their increasing popularity raises significant concerns. Addressing these biases is crucial to ensure fair and equitable outcomes.

\bibliographystyle{IEEEbib}
\bibliography{strings,refs}

\begin{appendices}
\section{Additional results}
In Table \ref{tab:BERTweet_demographic_breakdown} and Table \ref{tab:hateBERT_demographic_breakdown}, we assess positionality and model alignment for different demographics based on overall f1 score as well as TPR (recall). For data subsets and demographic identities associated with higher TPR and F1 scores, this suggests a model alignment with positionality.

\begin{table}[htbp]
    \centering
    \caption{Model performance breakdown for topic and demographic subsets for Toxigen\_HateBERT.}
    \label{tab:hateBERT_demographic_breakdown}
    \resizebox{0.7\columnwidth}{!}{
    \begin{tabular}{c c c c c}
    \toprule
        \textbf{Data Subset} & \textbf{Demographic} & \textbf{Micro F1} & \textbf{Precision} & \textbf{Recall} \\ 
        \midrule
        full & asian        & 0.5031   & 0.5031    & 0.5031  \\ \hline
        topic 0 & asian    & 0.5517   & 0.5517    & 0.5517  \\ \hline
        topic 1 & asian    & 0.4828   & 0.4828    & 0.4828  \\ \hline
        topic 2 & asian    & 0.4894   & 0.4894    & 0.4894  \\ 
        \midrule
        full & black       & 0.5385   & 0.5385    & 0.5385  \\ \hline
        topic 0 & black    & 0.3158   & 0.3158    & 0.3158  \\ \hline
        topic 1 & black    & 0.4286   & 0.4286    & 0.4286  \\ \hline
        topic 2 & black    & 0.4737   & 0.4737    & 0.4737  \\ 
        \midrule
        full & latino/latina & 0.5849 & 0.5849    & 0.5849  \\ \hline
        topic 0 & latino/latina & 0.7222 & 0.7222 & 0.7222  \\ \hline
        topic 1 & latino/latina & 0.55   & 0.55      & 0.55    \\ \hline
        topic 2 & latino/latina & 0.4667 & 0.4667    & 0.4667  \\ 
        \midrule
        full & man         & 0.4949   & 0.4949    & 0.4949  \\ \hline
        topic 0 & man      & 0.5526   & 0.5526    & 0.5526  \\ \hline
        topic 1 & man      & 0.5528   & 0.5528    & 0.5528  \\ \hline
        topic 2 & man      & 0.4348   & 0.4348    & 0.4348  \\ 
        \midrule
        full & native american & 0.6667 & 0.6667   & 0.6667  \\ \hline
        topic 0 & native american & 0.5  & 0.5      & 0.5     \\ \hline
        topic 1 & native american & 1.0 & 1.0      & 1.0     \\ \hline
        topic 2 & native american & 0.5  & 0.5      & 0.5     \\ 
        \midrule
        full & non-binary   & 0.4933   & 0.4933    & 0.4933  \\ \hline
        topic 0 & non-binary & 0.5455   & 0.5455    & 0.5455  \\ \hline
        topic 1 & non-binary & 0.4828   & 0.4828    & 0.4828  \\ \hline
        topic 2 & non-binary & 0.625    & 0.625     & 0.625   \\ 
        \midrule
        full & pacific islander & 0.7143 & 0.7143   & 0.7143  \\ \hline
        topic 0 & pacific islander & 0.3333 & 0.3333 & 0.3333 \\ \hline
        topic 1 & pacific islander & 1.0   & 1.0      & 1.0     \\ \hline
        topic 2 & pacific islander & 1.0   & 1.0      & 1.0     \\ 
        \midrule
        full & white       & 0.4857   & 0.4857    & 0.4857  \\ \hline
        topic 0 & white    & 0.5352   & 0.5352    & 0.5352  \\ \hline
        topic 1 & white    & 0.5309   & 0.5309    & 0.5309  \\ \hline
        topic 2 & white    & 0.4768   & 0.4768    & 0.4768  \\ 
        \midrule
        full & woman       & 0.5410   & 0.5410    & 0.5410  \\ \hline
        topic 0 & woman    & 0.4919   & 0.4919    & 0.4919  \\ \hline
        topic 1 & woman    & 0.5357   & 0.5357    & 0.5357  \\ \hline
        topic 2 & woman    & 0.5053   & 0.5053    & 0.5053  \\
        \bottomrule
    \end{tabular}
    }
\end{table}

\begin{table}[htbp]
    \centering
        \caption{Model performance breakdown for topic and demographic subsets for BERTweet.}
    \label{tab:BERTweet_demographic_breakdown}
    \resizebox{0.7\columnwidth}{!}{
    \begin{tabular}{c c c c c}
    \toprule

         \textbf{Data Subset} & \textbf{Demographic} & \textbf{Micro F1} & \textbf{Precision} & \textbf{Recall} \\  
            \midrule

        full & asian        & 0.5399   & 0.5399    & 0.5399  \\ \hline
        topic 0 & asian    & 0.5690   & 0.5690    & 0.5690  \\ \hline
        topic 1 & asian    & 0.5172   & 0.5172    & 0.5172  \\ \hline
        topic 2 & asian    & 0.5106   & 0.5106    & 0.5106  \\ 
        \midrule
        full & black       & 0.5962   & 0.5962    & 0.5962  \\ \hline
        topic 0 & black    & 0.3684   & 0.3684    & 0.3684  \\ \hline
        topic 1 & black    & 0.4286   & 0.4286    & 0.4286  \\ \hline
        topic 2 & black    & 0.5263   & 0.5263    & 0.5263  \\ 
        \midrule
        full & latino/latina & 0.5849 & 0.5849    & 0.5849  \\ \hline
        topic 0 & latino/latina & 0.6667 & 0.6667 & 0.6667  \\ \hline
        topic 1 & latino/latina & 0.6    & 0.6       & 0.6     \\ \hline
        topic 2 & latino/latina & 0.5333 & 0.5333    & 0.5333  \\ 
        \midrule
        full & man         & 0.5378   & 0.5378    & 0.5378  \\ \hline
        topic 0 & man      & 0.5395   & 0.5395    & 0.5395  \\ \hline
        topic 1 & man      & 0.5829   & 0.5829    & 0.5829  \\ \hline
        topic 2 & man      & 0.4275   & 0.4275    & 0.4275  \\ 
        \midrule
        full & native american & 0.75 & 0.75      & 0.75    \\ \hline
        topic 0 & native american & 0.5 & 0.5       & 0.5     \\ \hline
        topic 1 & native american & 1.0 & 1.0      & 1.0     \\ \hline
        topic 2 & native american & 0.5 & 0.5      & 0.5     \\ 
        \midrule
        full & non-binary   & 0.4933   & 0.4933    & 0.4933  \\ \hline
        topic 0 & non-binary & 0.4545   & 0.4545    & 0.4545  \\ \hline
        topic 1 & non-binary & 0.4828   & 0.4828    & 0.4828  \\ \hline
        topic 2 & non-binary & 0.6667   & 0.6667    & 0.6667  \\ 
        \midrule
        full & pacific islander & 0.5714 & 0.5714   & 0.5714  \\ \hline
        topic 0 & pacific islander & 0.3333 & 0.3333 & 0.3333 \\ \hline
        topic 1 & pacific islander & 1.0   & 1.0      & 1.0     \\ \hline
        topic 2 & pacific islander & 1.0   & 1.0      & 1.0     \\ 
        \midrule
        full & white       & 0.5165   & 0.5165    & 0.5165  \\ \hline
        topic 0 & white    & 0.5211   & 0.5211    & 0.5211  \\ \hline
        topic 1 & white    & 0.5494   & 0.5494    & 0.5494  \\ \hline
        topic 2 & white    & 0.4967   & 0.4967    & 0.4967  \\ 
        \midrule
        full & woman       & 0.5578   & 0.5578    & 0.5578  \\ \hline
        topic 0 & woman    & 0.5135   & 0.5135    & 0.5135  \\ \hline
        topic 1 & woman    & 0.5491   & 0.5491    & 0.5491  \\ \hline
        topic 2 & woman    & 0.5372   & 0.5372    & 0.5372  \\
    \bottomrule
    \end{tabular}
    }
\end{table}
\end{appendices}

\end{document}